\documentclass[journal]{IEEEtran}
\usepackage{amsmath,amsfonts}
\usepackage{algorithmic}
\usepackage{algorithm}
\usepackage{array}
\usepackage[caption=false,font=normalsize,labelfont=sf,textfont=sf]{subfig}
\usepackage{textcomp}
\usepackage{stfloats}
\usepackage{url}
\usepackage{amsmath}
\usepackage{verbatim}
\usepackage{graphicx}
\usepackage{cite}
\usepackage{titlesec}
\usepackage{booktabs}
\usepackage{multirow}
\usepackage{graphicx}
\usepackage{bbding}
\usepackage{CJKutf8}
\usepackage{amssymb}
\usepackage{amsfonts}
\begin{document}

\title{Deep Image Semantic Communication Model for\\ Artificial Intelligent Internet of Things}

\author{Li Ping Qian, \emph{Senior Member, IEEE}, Yi Zhang, Sikai Lyu, Huijie Zhu,\\
 Yuan Wu, \emph{Senior Member, IEEE}, Xuemin Sherman Shen, \emph{Fellow, IEEE}, and Xiaoniu Yang
        % <-this % stops a space
\thanks{Li Ping Qian, Yi Zhang and Sikai Lyu are with the College of Information Engineering, Zhejiang University of Technology, Hangzhou 310023, China (e-mail: lpqian@zjut.edu.cn).}
\thanks{Huijie Zhu is with the National Key Laboratory of Electromagnetic Space Security, Jiaxing 314033, China (e-mail: zhuhuijie@zju.edu.cn).}
\thanks{Yuan Wu is with the State Key Laboratory of Internet of Things for Smart City, University of Macau, Macau, China, with the Department of Computer Information Science, University of Macau, Macau, China, and also with the Zhuhai UM Science and Technology Research Institute, Zhuhai 519031, China (e-mail: yuanwu@um.edu.mo).}
\thanks{Xuemin Shen is with the Department of Electrical and Computer Engineering, University of Waterloo, Waterloo, ON N2L 3G1, Canada.
(e-mail: xshen@bbcr.uwaterloo.ca).}
\thanks{Xiaoniu Yang is with the Institute of Cyberspace Security, Zhejiang University of Technology, Hangzhou 310023, China, and also with the National Key Laboratory of Electromagnetic Space Security, Jiaxing 314033, China (e-mail: yxn2117@126.com).}}

% The paper headers
%\markboth{Journal of \LaTeX\ Class Files,~Vol.~14, No.~8, August~2021}
%{Shell \MakeLowercase{\textit{et al.}}: A Sample Article Using IEEEtran.cls for IEEE Journals}

%\IEEEpubid{0000--0000/00\$00.00~\copyright~2021 IEEE}
% Remember, if you use this you must call \IEEEpubidadjcol in the second
% column for its text to clear the IEEEpubid mark.

\maketitle

\begin{abstract}
With the rapid development of Artificial Intelligent Internet of Things (AIoT), the image data from AIoT devices has been witnessing the explosive increasing. In this paper, a novel deep image semantic communication model is proposed for the efficient image communication in AIoT. Particularly, at the transmitter side, a high-precision image semantic segmentation algorithm is proposed to extract the semantic information of the image to achieve significant compression of the image data. At the receiver side, a semantic image restoration algorithm based on Generative Adversarial Network (GAN) is proposed to convert the semantic image to a real scene image with detailed information. Simulation results demonstrate that the proposed image semantic communication model can improve the image compression ratio and recovery accuracy by 71.93\%  and 25.07\%  on average in comparison with WebP and CycleGAN, respectively. More importantly, our demo experiment shows that the proposed model reduces the total delay by 95.26\%  in the image communication, when comparing with the original image transmission.
\end{abstract}


\section{Introduction}
\IEEEPARstart{W}{ith} the significant development of Internet of Things 
(IoT) and Artificial Intelligence (AI), the AIoT industry 
has undergone rapid growth recently\cite{cite-key}. The world is transitioning from an era of connected everything to an era of intelligent connected everything. AIoT combines the strengths of artificial intelligence and IoT to enable devices to more effectively analyze and understand a wide range of data collected, including text, voice, and images\cite{9061155}. IoT technology improves business capabilities and service levels for AI through de-vice connectivity, signaling, and data exchange. AIoT has applications in many fields, such as smart factories that improve the productivity of industrial robots by embedding sensors, and automobiles that rely on fast IoT applications for autonomous driving\cite{6337127}. These applications deeply integrate smart computing and communication technologies, allowing communication scenarios to shift from traditional human-to-human and human-to-object communication to communication between intelligences. According to statistics, AIoT devices generate approximately 10GB of data per day. By 2030, it is expected that 350 billion devices will be connected to the global IoT\cite{9062950}, so the amount of data to be collected, stored, and analyzed will grow exponentially. However, due to the limitations of cloud computing capabilities, traditional cloud computing methods are unable to eliminate redundant information when dealing with the large amount of data transmitted by AIoT devices\cite{8276823}, and cannot meet the requirements of low-latency and high-precision tasks. In addition, tight bandwidth resources limit the efficiency of data transmission between IoT devices and the cloud\cite{ZAHOOR2021921}, making the task of transmitting all data from IoT devices extremely difficult. Therefore, efficient data compression and transmission have become important challenges nowadays\cite{10163877}.

As a cutting-edge intelligent communication method, semantic communication is dedicated to extracting data from the semantic level and filtering out irrelevant and redundant information\cite{10007890}. Its core objective has shifted from the accuracy of transmitted data and signal waveforms to ensuring the accurate understanding and correct use of the transmitted semantic information\cite{2323}. Therefore, semantic communication can minimize data redundancy, achieve effective data compression, and improve bandwidth utilization under the premise of ensuring high-level task performance. This not only reduces the waste of spectrum resources and lowers communication energy consumption, but also better meets human needs and intentions. Additionally, low-capacity semantic communication remains robust in harsh channel environments (i.e., low signal-to-noise ratio regions), making it ideal for AIoT applications that require high reliability and low-latency.

With the rapid development of Deep Learning (DL) and hardware devices, semantic communication based on deep learning has gradually received widespread attention \cite{9360873}.In text semantic communication, Xie et al.\cite{9252948} proposed a deep learning-based distributed semantic communication system, which realizes the semantic transmission of data from the IoT terminal to the cloud/edge as well as the improvement of efficiency. Qian et al.\cite{99999999} proposed a semantic extraction model for Chinese text which improves the quality of semantic extraction by integrating the Transformer with the Pointer-Generator NetworkIn terms of semantic communication for voice data, in order to realize high-precision audio transmission with a small amount of data, Tong et al. \cite{9685654} in estigated the problem of audio-based semantic communication in wireless networks, and proposed a self-encoder consisting of a Convolutional Neural Network (CNN) based on a wave vector architecture. Nowadays, AIoT improves human-computer interaction patterns and enhances data management and analysis by understanding and analyzing data such as text and images through artificial intelligence techniques such as deep learning. Image data plays an important role as a medium and means to convey information and is widely used. Therefore, image semantic communication systems based on deep learning have been gradually developed. Huang et al.\cite{9685667} proposed a method to use GAN for image semantic coding, which constructs a semantic coding model from coarse to fine for multimedia semantic communication systems. To maximize the analysis performance of target-based correlation tasks, Patwa et al. \cite{9191247} proposed a method to compress visual data. Sun et al. \cite{unknown} proposed a pixel semantics-based joint source channel coding method that maintains the semantic information behind pixels when images are transmitted without boundaries. Wang et al. \cite{10001359} introduced adversarial loss into deep learning-based joint channel coding in which it is optimized. The algorithm can better preserve the local texture features of an image while maintaining the overall semantics of the image. Wang \cite{9126173} achieved adaptive enhancement of semantic features by constructing a semantic a priori attention model. Hu et al. \cite{10012843} proposed a framework for an end-to-end semantic communication system that significantly enhances the robustness of the semantic communication system against semantic noise while reducing the transmission overhead.

Although existing image semantic communication models can provide reasonably good semantic communication results, there are still three areas that can be further improved:

\begin{itemize}
\item{Some of the existing deep learning-based image semantic segmentation algorithms still suffer from the problem of inaccurate segmentation accuracy or the inability to effectively balance segmentation accuracy and compression performance.}
\item{Most image semantic communication models do not account for the importance of semantic restoration of semantic images to real images, which leads to the inability to obtain original texture and detail information from the image to cope with corresponding requirements, and affects the viewability of the image.}
\item{As the depth of the network continues to increase, the amount of storage and computation in the network grows exponentially, and the insufficiently lightweight network models limit their feasibility for specific applications in AIoT.}
\end{itemize}

To address the aforementioned issues, this paper proposes a deep image semantic communication model for AIoT, and the main research contributions are as follows.

\begin{itemize}
\item{At the transmitter side of the communication system, a high-precision image semantic segmentation algorithm is proposed. Initial feature extraction of the image is carried out by a residual network, and the pyramid pooling and attention mechanism are introduced to further enhance feature information. This approach achieves significant compression of image data while extracting high-quality semantic information from the image.}
\item{At the receiver side of the communication system, a semantic communication restoration algorithm based on GAN is proposed to reduce semantic communication to real scene images with rich detail information. This not only improves the communication system but also provides sustainability for subsequent downstream tasks involving the image.}
\item{The semantic segmentation model was quantized in order to reduce the computational memory of the model, speed up the inference of the model, and lower the storage memory requirements. By using integers instead of floating point for storage and computation, the model footprint is reduced by a factor of 4, which plays a significant role in promoting the practical application of AIoT.}
\end{itemize}
\section{system model}
As shown in Fig.1, we proposes an image semantic communication system which mainly consists of three components: the transmitter, the channel, and the receiver. The transmitter's function is primarily to generate a semantic image by performing semantic segmentation on the original image. The channel is responsible for transmitting the information, while the receiver's function is mainly to reduce the semantic image into a reconstructed image with rich detail information that is visually realistic. 

\begin{figure}[ht]
\centering
\includegraphics[width=3.35in]{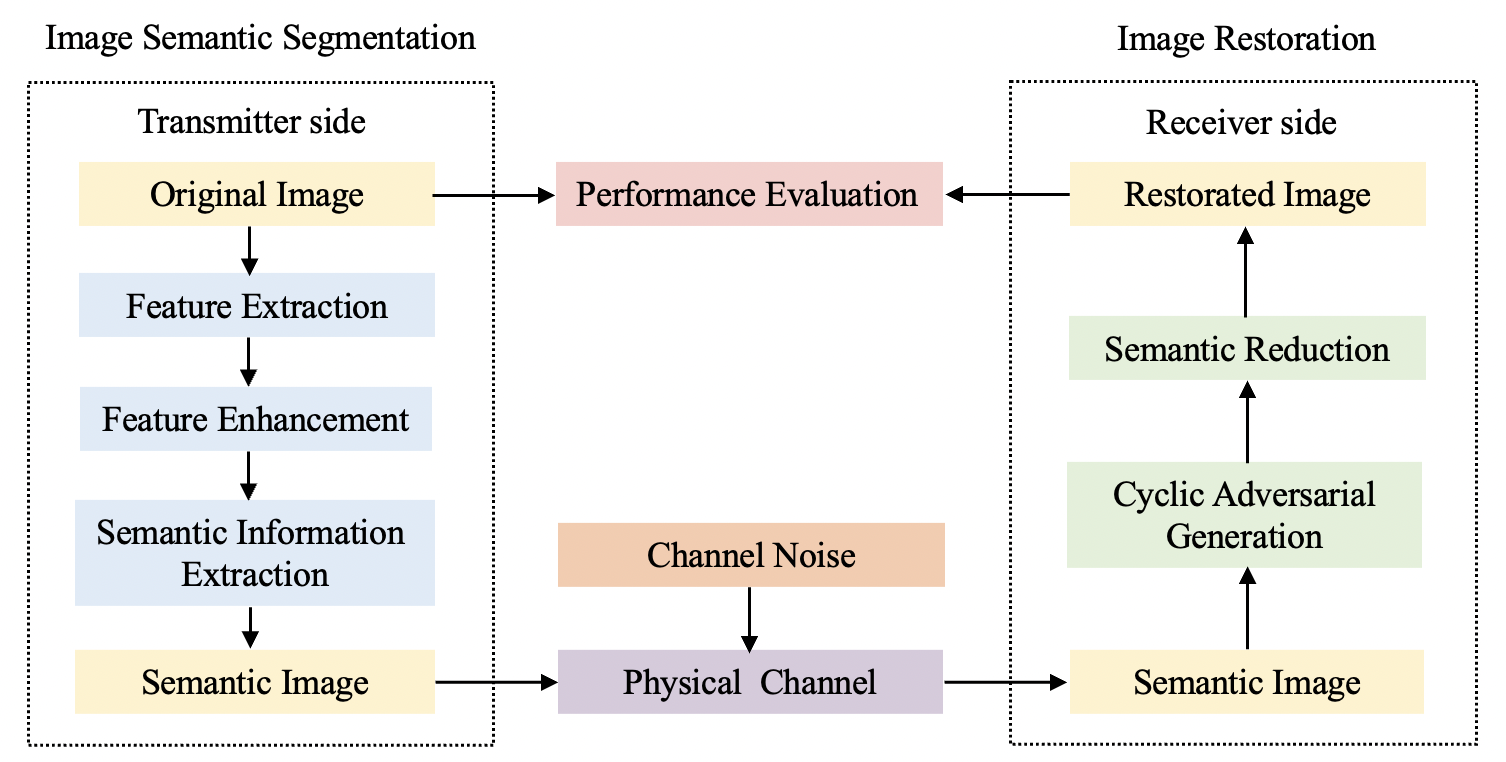}
\caption{Schematic Diagram of Proposed Deep Image Semantic Communication System Model.}
\label{fig_1}
\end{figure}

\subsection{Problem Description}
The objective of the proposed image semantic communication system is to acquire the semantic information of the original image as accurately as possible at the transmitter side to generate a semantic image. While at the receiver side, it is to reconstruct an image that effectively restores the semantic features and detail information of the original image. The main change from the conventional approach is in the data processing phase before transmitting and after receiving\cite{9955312}.The design of the model mainly involves two aspects. One is how to improve the accuracy of semantic segmentation while maintaining the compression performance of the operation. The second is how to restore the detailed features of the semantic image more effectively, while ensuring visual realism.

\subsection{Image Semantic Segmentation and Image Restoration}
As shown in Fig. 1, the transimitter side consists of three modules: feature extraction, feature enhancement and semantic information extraction. Among them, the feature extraction module utilizes the CNN as a feature extraction network to extract the primary features from the original image \textit{m}. The feature enhancement module takes the primary features of the image and enhances the expression of the features by combining the location information with the semantic information
%figure 2
\begin{figure*}[t]
\centering
\includegraphics[width=6.6in]{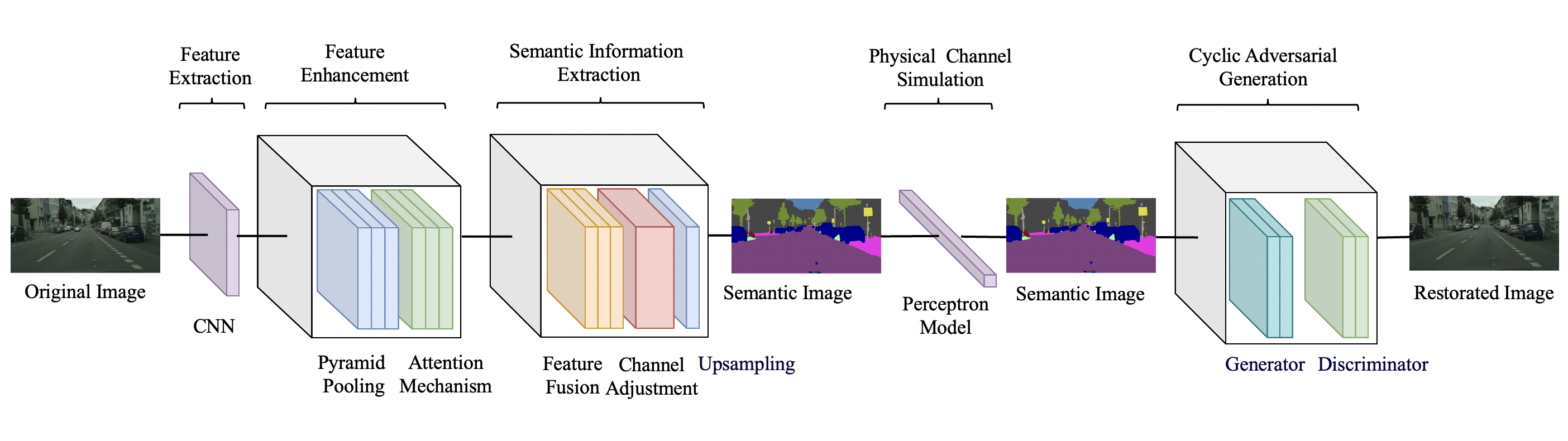}
\caption{Framework of Proposed Deep Image Semantic Communication System Model.}
\label{fig_2}
\end{figure*}
through the pyramid pooling and the attention mechanism. Finally, the semantic information extraction extracts semantic information from the feature map through three steps of feature fusion, channel adjustment, and upsampling, assigns the corresponding label to each pixel, and outputs the semantic image. Therefore, the transmitting signal at the transimitter side can be expressed as
\begin{equation}
\label{deqn_ex1a1}
\ x=S\left(m,\theta\right),
\end{equation}
where \textit{x} is the semantic image obtained by semantic segmentation of the original image; \textit{S(·)} is the process of semantic segmentation of the image; $\theta$ is the set of parameters of semantic segmentation of the image. The transimitter side sends \textit{x} out and reaches the receiver side through the physical channel, and the signal at the receiver side which is affected by the channel noise can be represented as
\begin{equation}
\label{deqn_ex1a2}
\ y=hx+n,
\end{equation}
where \textit{y} is the semantic image received at the receiver side; \textit{h} is the channel gain of the Rayleigh fading channel; and \textit{n} is the additive white Gaussian noise (AWGN). For end-to-end training of encoders and decoders, the physical channel must allow for backpropagation. Therefore, the physical channel can be modelled using a neural network. 

On the other hand, the receiver side utilises the principle of cyclic adversarial generation to reconstruct the semantic reduced image into a detail-rich restored image that is visually realistic. As a result, the restored image can be represented as
\begin{equation}
\label{deqn_ex1a3}
\ m^\prime=S^{-1}\left(y,\delta\right),
\end{equation}
where $m^\prime$ is the reduced image; \textit{$S^{-1}$}\textit{(·)} is the semantic restoration process; $\delta$ is the set of parameters of the semantic restoration process. The goal of this paper is to reconstruct an image as similar as possible to the original image, therefore the mean square error (MSE) can be used as the objective function of the image semantic model, i.e.,
\begin{equation}
\label{deqn_ex1a4}
L_{MSE}\left(m,m^\prime\right)=\mathop{\min}\limits_{\theta,\delta}{\left(m,m^\prime\right)^2}.
\end{equation}

%section deep image semantic communication model
\section{deep image semantic communication model}
\subsection{System Design and Operation Process}
As shown in Fig. 2, the model design is mainly divided into three parts: transimitter side, channel and receiver side.
\begin{itemize}
\item{Transmitter side. }
The transmitter side is used to realize the entire process of semantic segmentation and send the semantic image to the channel. Specifically, the original image is input into the transmitter side, and the semantic image is obtained after feature extraction, feature enhancement, and semantic information extraction in turn. The goal of the transmitter side is to realize efficient feature extraction of semantic information and semantic image output, so that the semantic image is more accurate and efficiently compressed.
\item{Channel. }
The channel is used to transmit the semantic image formed at the transmittor side to the receiver side to realize efficient transmission with high accuracy and low-latency.
\item{Receiver side. }
The receiver side is used to reduce the received semantic image to a real scene image rich in detail information using the principle of cyclic adversarial generation.
\end{itemize}

\subsection{Image Semantic Segmentation}
%figure 3
\begin{figure}[ht]
\centering
\includegraphics[width=3.4in]{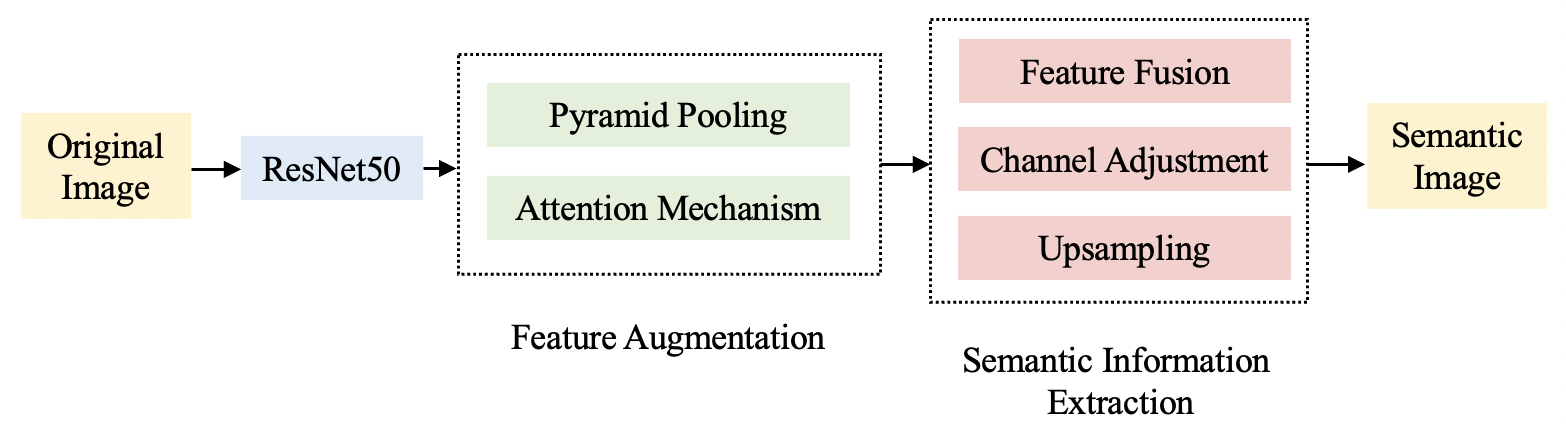}
\caption{Framework of Image Semantic Segmentation Model.}
\label{fig_3}
\end{figure}

As shown in Fig.3, the process of image semantic segmentation mainly includes three modules: feature extraction, feature enhancement, and semantic information extraction. The basic process is as follows.
\begin{itemize}
\item{The original image is taken as input and primary features are extracted by ResNet-50 [15];}
\item{The primary features output from ResNet-50 are fed into the pyramid pooling module, which consists of a four-layer structure. It divides the feature map into four feature maps, which are in turn divided into 1×1, 2×2, 3×3, and 6×6 sub-regions. An average pooling operation is performed on each subregion within the four feature maps separately, and subsequently these four feature maps are stitched together. The dimensionality of the feature maps is reduced to 1/8 of the size of the original feature maps using a convolutional kernel, and then upsampling operations are performed on these four low-dimensional features to restore the dimensions of the feature maps to the size of the original features. Finally, these feature maps obtained from upsampling are stitched together with the original features to form the output feature maps of the pyramid pooling module;}
\item{The above output feature map is used as input to the attention mechanism module to extract high-level features of the image through the spatial attention mechanism module and the channel attention mechanism module, respectively;}
\item{Using the image features obtained in the above steps, the image semantic information is extracted using three steps: feature fusion, channel adjustment, and upsampling.}
\end{itemize}

In the following,we would introduce the operation of each module in details.
\subsubsection{Feature Extraction}

In traditional network models, as the depth of the network model increases, the number of network parameters also increases, while the feature extraction capability is also enhanced. However, when increasing to a certain level, problems such as gradient vanishing or gradient explosion occur\cite{9159868}. In order to solve these problems, researchers have proposed methods to optimize the input and intermediate layer data through normalization operations \cite{article}. However, during the training process, extremely deep networks may suffer from the degradation problem, i.e., the phenomenon that the accuracy instead decreases as the depth of the network increases.The emergence of ResNet-50 effectively solves the above problems \cite{7780459}.

\begin{table}[ht]
\caption{ResNet Network Architecture\label{tab:tabel1}}
\centering
\begin{tabular}{@{}c|c|c@{}}
\toprule
Layer name                & Net                                                                                & Output                       \\ \midrule
Conv1                     & \begin{tabular}[c]{@{}c@{}}7×7,64,stride 2,Batch\\ Normalization,ReLU\end{tabular} & H/2×W/2×64                   \\ \midrule
\multirow{2}{*}{Conv2\_x} & 3×3,Maxpooling,stride 2                                                            & \multirow{2}{*}{H/4×W/4×256} \\ \cmidrule(lr){2-2}
                          & $\left[\begin{matrix}1\times1,&64\\3\times3,&64\\1\times1,&256\\\end{matrix}\right]\times3$                                                                               &                              \\ \midrule
Conv3\_x                  & $\left[\begin{matrix}1\times1,&128\\3\times3,&128\\1\times1,&512\\\end{matrix}\right]\times4$                                                                               & $H/4×W/4×512$                  \\ \midrule
Conv4\_x                  & $\left[\begin{matrix}1\times1,&256\\3\times3,&256\\1\times1,&1024\\\end{matrix}\right]\times6$                                                                               & $H/8×W/8×1024$                 \\ \midrule
Conv5\_x                  & $\left[\begin{matrix}1\times1,&512\\3\times3,&512\\1\times1,&2048\\\end{matrix}\right]\times3$                                                                               & $H/8×W/8×2048$                 \\ \bottomrule
\end{tabular}
\end{table}

ResNet-50 is mainly composed of a constant residual block (Identity block) and a convolutional residual block (Conv block). The Conv block consists of two convolutional layers, which can transform the size of the input feature maps so that the size of the output feature maps matches the size of the input feature maps of the subsequent Identity block. The Identity block consists of three convolutional layers, which realize constant mapping, allowing the network structure to deepen while avoiding the addition of huge network parameters, so that the neural network remains optimal even when the depth is increased, and network performance does not degrade. Therefore, ResNet-50 can improve the quality of feature extraction by increasing the depth of the network by successively accessing multiple constant residual blocks after the Conv blocks.

The feature extraction module utilizes Res-Net50 to extract primary features, and its network structure is summarized in Table 1. The network is divided into five stages, beginning with the input image of size H×W×3. The first stage involves a convolution operation using 64 convolution kernels with a stride of 2 and a kernel size of 7×7, resulting in an output feature map of size H/2×W/2×64. The second stage firstly performs a max-pooling operation using a stride of 2 and a kernel size of 3×3, and then connects three repeated residual structures. Each residual structure contains three convolution kernels of sizes 1×1, 3×3, and 1×1 to extract features, as detailed in Table 1. The output feature map size of this stage is H/4×W/4×256. The third stage involves four connected residual structures, with an output feature map size of H/8×W/8×512. The fourth stage involves six connected residual structures, with an output feature map size of H/8×W/8×1024. Finally, the fifth stage connects three residual structures and the output feature map size becomes H/8×W/8×2048.

\subsubsection{Feature Enhancement}
In image semantic segmentation, the contextual information of an image affects the segmentation effect\cite{8954873}. Pyramid Pooling\cite{7005506} improves the segmentation effect of the network model by extracting global contextual semantic information.

\begin{figure}[ht]
\centering
\includegraphics[width=3.4in]{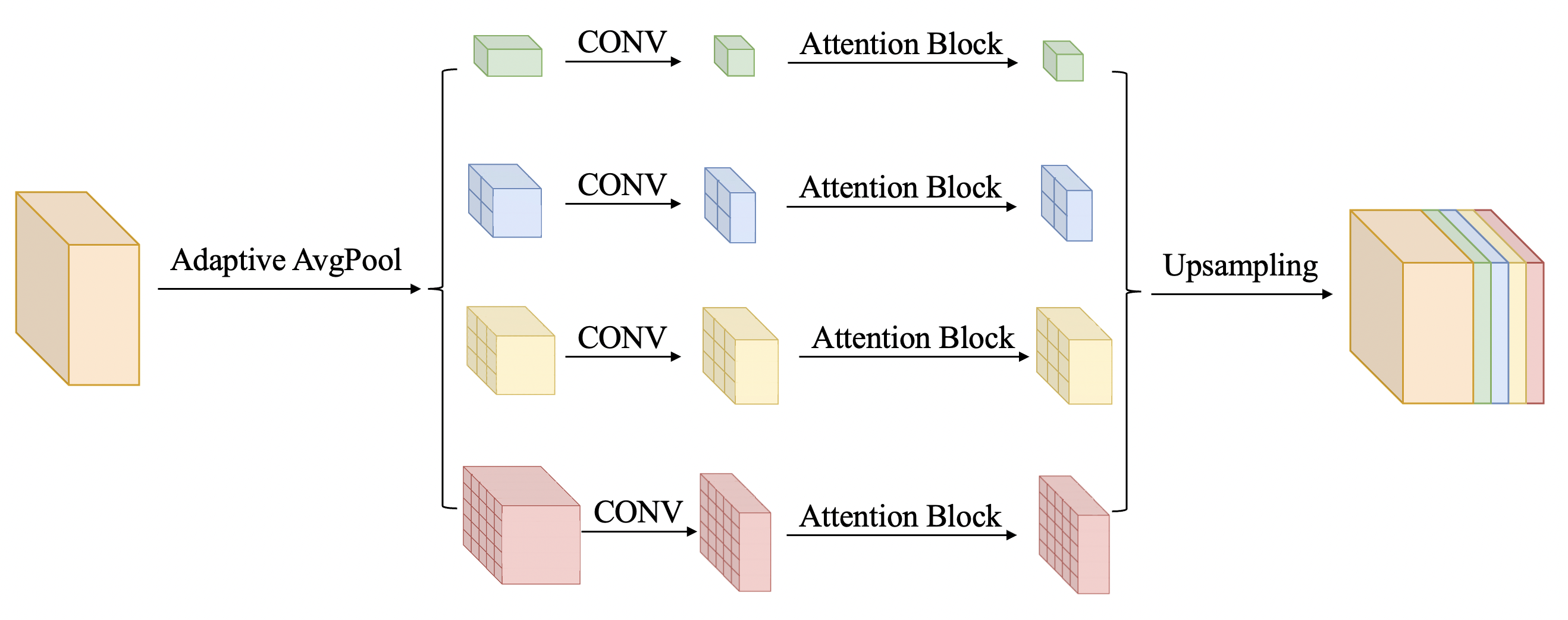}
\caption{Schematic Diagram of Pyramid Pooling.}
\label{fig_4}
\end{figure}
As shown in Fig.4, the module consists of four layers. The first layer only has one 1×1 region and performs global average pooling to obtain a feature map of size 1×1×2048. The second layer divides the feature map into four 2×2 subregions and performs average pooling within each subregion to obtain a feature map of size 2×2×2048. The third layer divides the feature map into 3×3 subregions(9 in tatal) and performs average pooling within each subregion to obtain a feature map of size 3×3×2048. The fourth layer divides the feature map into 6×6
subregions(36 in total) and performs average pooling within each subregion to obtain a feature map of size 6×6×2048. Subsequently, a convolution operation is performed on each subregion using a convolution kernel of size 1×1, which reduces the number of channels to 1/4 of the original, i.e.,then, the attention mechanism is used to increase the expression of features for each feature map. Subsequently, bilinear interpolation is used to upsample each feature map, and the dimensions of individual feature maps are restored to the dimensions prior to inputting into the pyramid pooling module and are compared with the original feature maps. Finally, the feature maps are spliced together to obtain a feature map of dimensions H/8×W/8×4096. In this way, contextual features from multi-layer images can be fused to obtain global image features.

\begin{figure}[ht]
\centering
\includegraphics[width=3.4in]{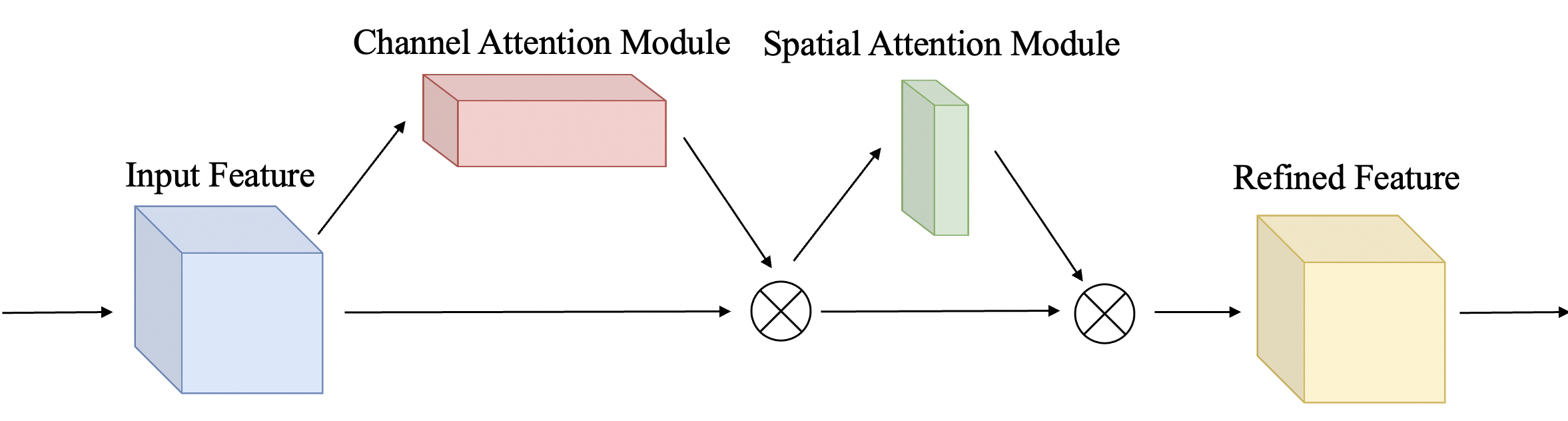}
\caption{Schematic Diagram of Attention Mechanism Module.}
\label{fig_5}
\end{figure}
\begin{figure}[ht]
\centering
\includegraphics[width=3.3in]{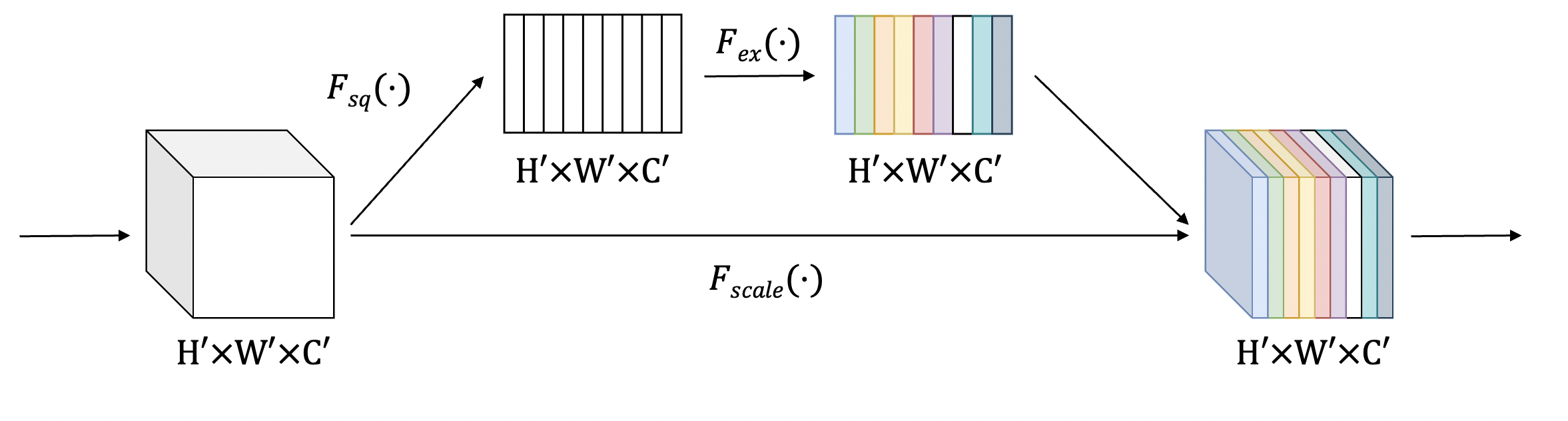}
\caption{Schematic Diagram of Channel Attention Mechanism.}
\label{fig_6}
\end{figure}
\begin{figure}[ht]
\centering
\includegraphics[width=3.4in]{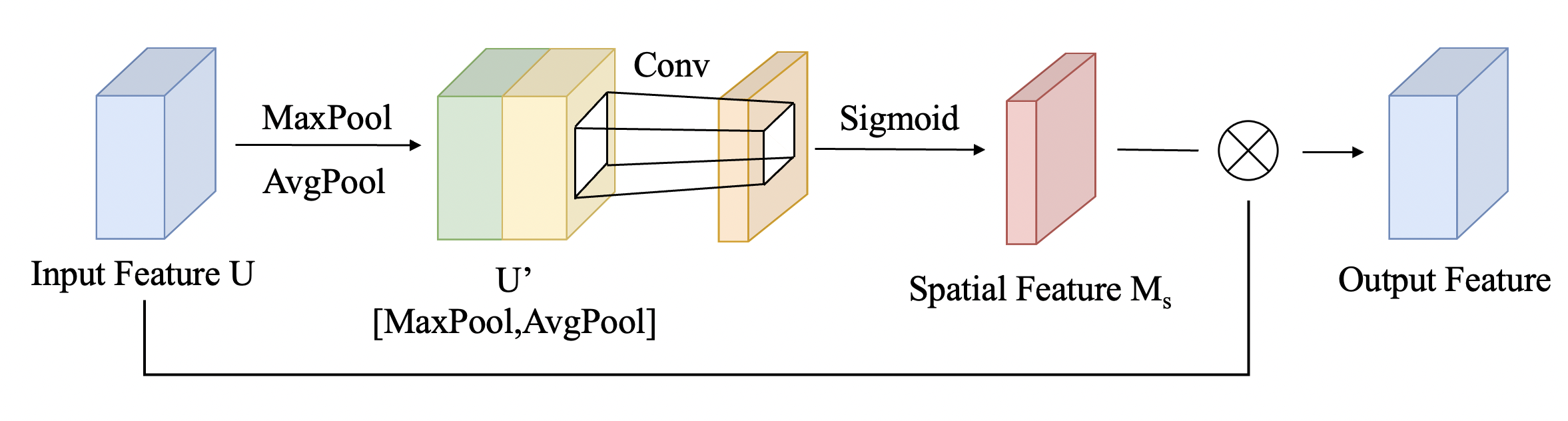}
\caption{Schematic Diagram of Spatial Attention Mechanism.}
\label{fig_7}
\end{figure}
The Attention Block was initially applied in the field of natural language processing, and has since been widely used in computer vision and other fields due to its ability to effectively capture the region that requires focused attention. The attention mechanism allows a computer to extract key information of interest from a large amount of information. Neural network determine the key regions of interest through the weight values of each feature in each image, and suppress the unimportant parts by reducing their weights\cite{2017Attention}. The attention mechanisms incorporated into this algorithm include the Channel Attention Module and the Spatial Attention Module, as shown in Fig.5. 

The first component of the attention mechanism module introduced in this paper is the channel attention mechanism. It can be seen as an attention operation in the feature space that helps the network to focus on important parts of the input and to reduce the influence of irrelevant information. The main idea is to weight the response of each position in the feature map according to the importance of each channel in the input feature map. This is achieved by learning a weight vector that allows the network to adaptively choose which channels are more important for a given task. As shown in Fig.6,\textit{ H'}, \textit{W'}, and \textit{C'} in the figure represent the dimensions of the input feature map for the attention mechanism, respectively. Specifically, the channel attention mechanism consists of compression, excitation, and weighting operations. The compression operation is $F_{sq}$ and calculates the distribution of features within each channel in the input feature map by performing a global average pooling operation on the input features. The calculation formula is
\begin{equation}
\label{deqn_ex1a5}
Z_C=F_{sq}\left(U_c\right)=\frac{1}{H\prime\times W\prime}\sum_{i=1}^{H}\sum_{j=1}^{W}{U_C\left(i,j\right)},
\end{equation}
where $Z_C$ is the output feature after the compression operation; $U_c$ is the feature of channel \textit{C} of the input feature map. The input image is squeezed by the squeezing operation to obtain a \textit{1×1×C} feature map. The excitation operation is $F_{ex}$ in Fig.6. It is performed by the Sigmoid function, which is used to capture the feature dependencies between each channel. The weighting operation is $F_{scale}$ in Fig.6. The output features are obtained by multiplying the channel weights obtained from the excitation operation and the input feature maps in the channel direction, which can multiply the input features by the corresponding weights respectively to get the corresponding weight values, and the weights of the channel features are adjusted according to the data after the input, and finally, the feature maps with the same size as the input size are obtained.

The second component is the spatial attention mechanism, similar to the channel attention mechanism.It is used to select important regions or locations in the input feature map and give them more weight for better feature extraction. The core idea is to focus on regions of interest by performing a weighted average over the spatial dimensions of the input feature map. In traditional CNN, each convolutional kernel slides over the entire feature map, while the spatial attention mechanism calculates weights for each location, enabling the model to automatically learn which locations should be given higher importance. Specifically, as shown in Fig.7, the spatial attention mechanism performs local max-pooling and local average pooling operations over all channel ranges for each feature point in the input feature map \textit{U} to obtain the maximum and mean values. Next, the results of these two computations are stacked and spliced to obtain the feature map \textit{U’}. Then, a convolution kernel is used to adjust the number of channels to 1, and then a Sigmoid function is applied to obtain a weight value ranging between [0,1] for each feature point, and this weight value is multiplied with the original input layer to obtain a feature map of the same dimensions as the one with the input feature map that contains the weighting of the spatial attention mechanism. In this way, the model can enhance the feature information that needs to be emphasized while suppressing redundant and useless feature information.

\subsubsection{Semantic Information Extraction}
After the pyramid pooling module and the advanced semantic feature extraction module are fused with the attention mechanism, the model successfully extracts the semantic features of the image. Next, the semantic information extraction module will be utilized to output the semantic information of the image. As shown in Fig.8, the module is divided into three steps in total:
\begin{itemize}
\item{Feature Fusion}

This phase integrates the features using a convolutional kernel of size 3×3 to obtain a more comprehensive representation of the features.
\item{Channel Adjustment}

This phase uses a convolutional kernel of size 1×1 to adjust the number of channels to a preset number of target object classes, helping assign target objects to specific pixels.
\item{Upsampling}

This phase aligns the width and height of the output layer with the input image by upsampling, ensuring that the output semantic information matches the size of the input image.
\end{itemize}
\begin{figure}[ht]
\centering
\includegraphics[width=3.4in]{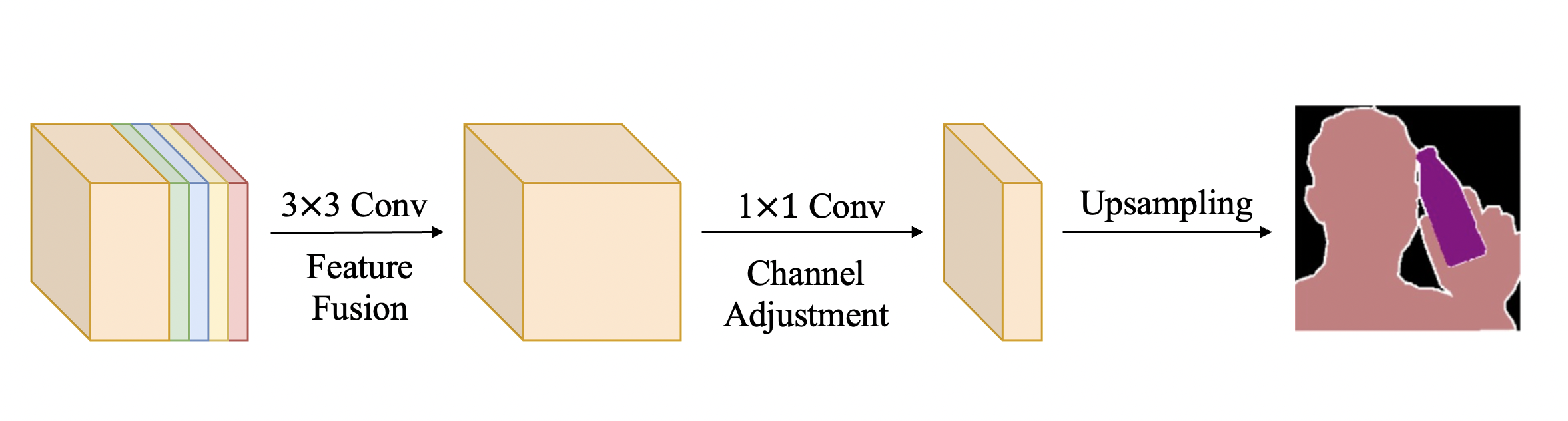}
\caption{Schematic Diagram of Semantic Information Extraction.}
\label{fig_8}
\end{figure}

\subsubsection{Loss Function}
To enhance the segmentation performance of the semantic segmentation model, the loss function of this algorithm mainly consists of the Dice Loss\cite{li2020dice}, the Focal Loss\cite{8417976}, and the Cross-Entropy Loss\cite{8693661}.
\begin{itemize}
\item{Cross-Entropy Loss}

Cross-Entropy loss is used to quantify the discrepancy between the model's prediction of the true label and the actual label, i.e., to assess the degree of similarity between the model's predicted probability distribution and the actual probability distribution. The formula for this loss is given as
\begin{equation}
\label{deqn_ex1a6}
\ L=\frac{1}{N}\sum_{i}{L_i=-\frac{1}{N}\sum_{i}\sum_{c=1}^{M}{y_{ic}log{\left(p_{ic}\right)}}}
\end{equation}
where \textit{N} is the size of the data set; \textit{M} is the number of categories; $y_{ic}$ is the sign function (0 or 1), if the real category of sample i is equal to c take 1, otherwise take 0; $p_{ic}$ is the predicted probability of observing that sample \textit{i} belongs to the category \textit{c}. When the probability distribution predicted by the model is exactly the same as the real label, the Cross-Entropy Loss attains the minimum value of 0. As the gap between the probability distribution predicted by the model and the real label increases, the Cross-Entropy Loss rises. Therefore, minimizing the Cross-Entropy Loss during training enables the model to make more accurate predictions for different categories.
\item{Dice Loss}

Dice loss is based on Dice coefficient to calculate similarity. The formula is calculated as
\begin{equation}
\label{deqn_ex1a7}
dice\ loss=1-dice=1-2\frac{\left|X\cap Y\right|}{\left|X\right|+\left|Y\right|},
\end{equation}
where \textit{X} is the model predicted outcome and \textit{Y} is the true outcome.
\item{The Focal Loss}

The Focal Loss is used to solve the problem of sample non-equilibrium while contributing to the overall improvement of the model performance. The formula is calculated as
\begin{equation}
\label{deqn_ex1a8}
 Focal\ Loss(p_t)=-\alpha_t{(1-p_t)}^\gamma log(p_t),
\end{equation}
where $\alpha_t$ is the category weight, which is used to regulate the proportion of positive and negative samples in the network;$p_t$ is the categorization probability of the recognized object; $\gamma$ is the focusing parameter; and ${(1-p_t)}^\gamma$ constitutes the sample difficulty weight adjustment factor.
\end{itemize}

\subsubsection{Model Quantification}
With the widespread adoption of DL in various fields such as computer vision and natural language processing, a plethora of deep learning-based networks have emerged. These models are characterized by their large size and complexity, which makes them suitable for inference on N-cards but not practical for deployment in embedded devices such as mobile phones\cite{9983851}. Additionally, customers often require deployment of these complex models in low-cost embedded devices, posing a significant challenge. To effectively address this challenge, model quantization techniques have emerged. These techniques enable compression of models with only a small sacrifice in accuracy, paving the way for applying these complex models to embedded terminals such as mobile phones and robots.

\begin{figure}[ht]
\centering
\includegraphics[width=2.2in]{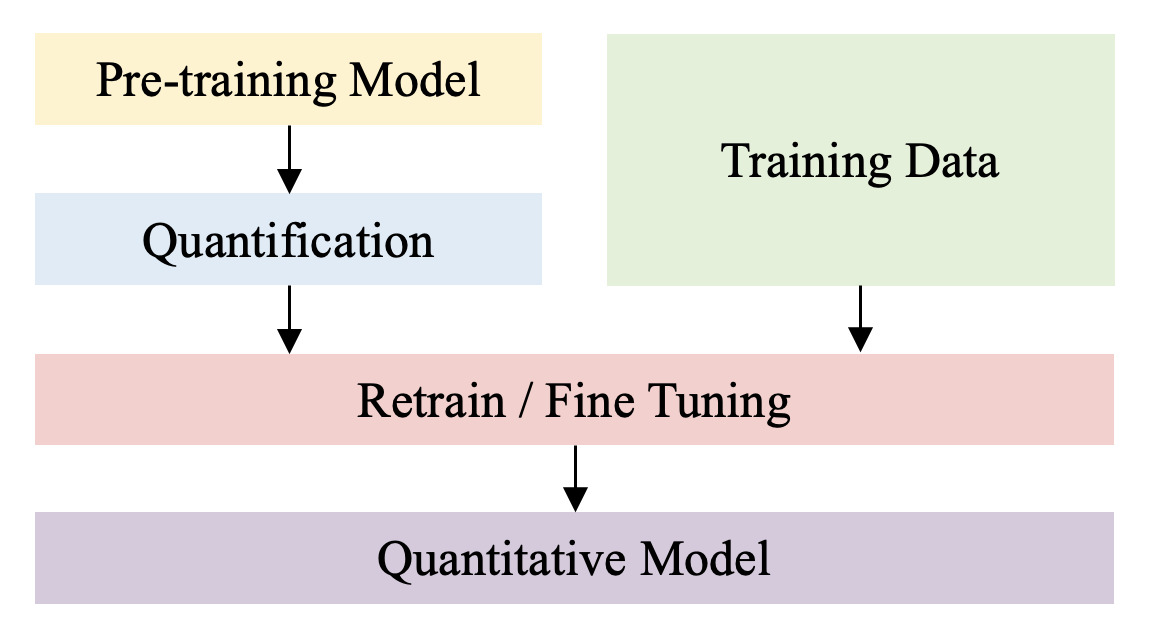}
\caption{Schematic Diagram of Quantification Module.}
\label{fig_9}
\end{figure}
With the advancement of model prediction accuracy and the increase in network depth, the amount of memory consumed by neural networks has become a paramount concern, especially for mobile devices. Typically, current mobile phones are equipped with 4GB of RAM to support the simultaneous operation of multiple applications, whereas three models running once typically occupy 1GB of RAM. Model size is not only a matter of memory capacity but also a concern for memory bandwidth. The models utilize model weights for each prediction, and image-related applications typically require real-time processing of data, which entails at least 30 FPS.

To minimize storage overhead and bandwidth requirements while achieving faster computation, we quantize the trained semantic segmentation model. We use 8-bit integers instead of 32-bit floating point numbers for storage and computation, thereby reducing the model storage footprint by a factor of 4. The process of model quantization is depicted in Fig.9.

\subsection{Image Semantic Restoration}
After extracting the semantic information from the image, performing semantic restoration on the original image at the receiver side is crucial for the construction and improvement of communication systems. The specific algorithm flow is shown in Fig.10. Assuming that the data in the \textit{X} domain and \textit{Y} domain are ${\{X_i\}}_{i=1}^N$  and ${\{Y_i\}}_{i=1}^M$, respectively, and the distributions of the data are $x-{P_{data}(x)}$ and $y-P_{data}(y)$, the image \textit{x} in the \textit{X} domain is mapped by the generator \textit{G} into the image \textit{G(x)} in the \textit{Y} domain. The discriminator corresponding to the generator \textit{G} is denoted as $D_Y$, which distinguishes whether the image \textit{G(x)} is real data or generated data. 

\begin{figure}[ht]
\centering
\includegraphics[width=3.4in]{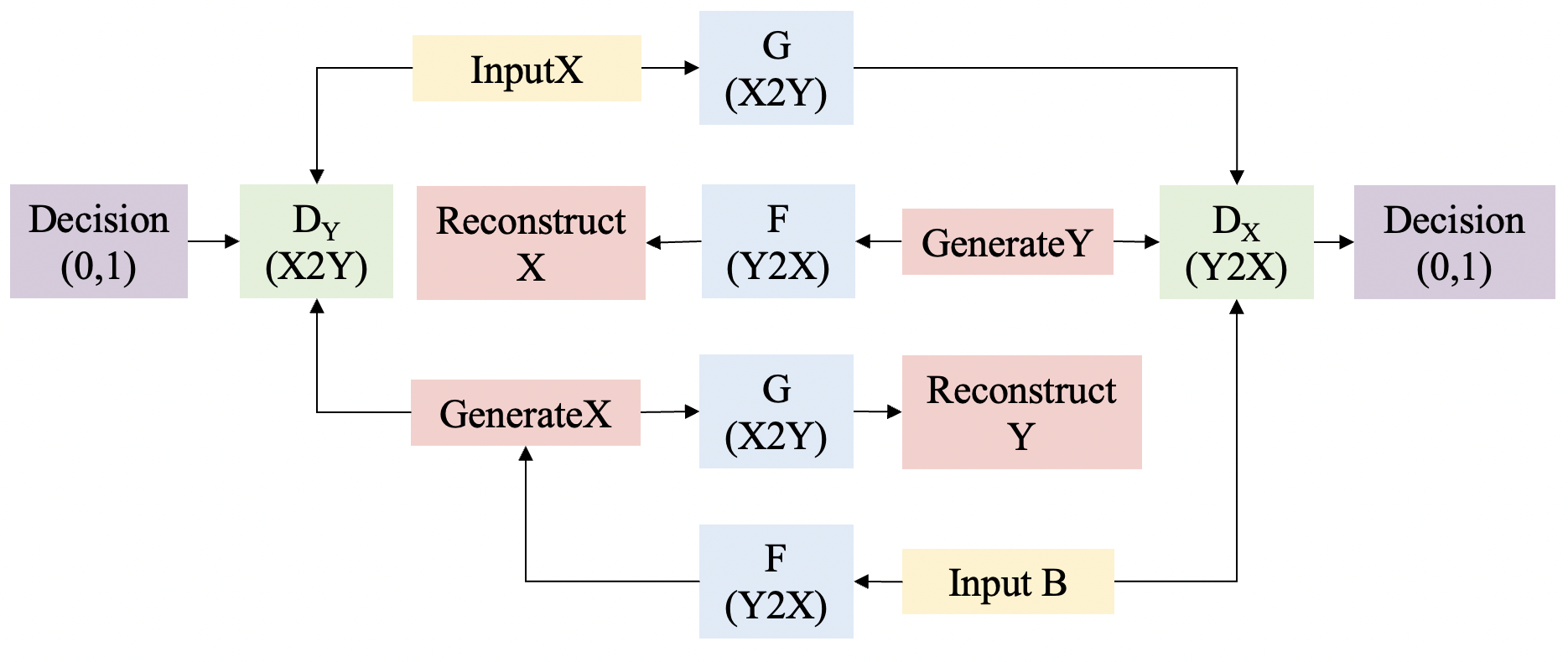}
\caption{Schematic Diagram of Image Semantic Restoration Algorithm.}
\label{fig_10}
\end{figure}
Since the above constitutes a single generative adversarial process, it cannot achieve the desired training effect. To avoid the situation where the generator \textit{G} can easily fool the discriminator $D_Y$ if it maps all the data in the \textit{X} domain to the same image in the \textit{Y} domain, based on the idea of pairwise learning, this algorithm adds cycle consistency loss as a constraint. Additionally, another generator \textit{F} is introduced to establish a mapping relationship from the \textit{Y} domain to the \textit{X} domain. After mapping \textit{F}, the image \textit{y} in the \textit{Y} domain is mapped to become the image \textit{F(y)} in the \textit{X} domain. The discriminator corresponding to the generator \textit{F} is denoted as $D_X$, which distinguishes whether the image \textit{F(y)} is real data or generated data.

Therefore, the model must learn both the \textit{G} and the \textit{F} mapping relationship during the training process. Meanwhile, it needs to satisfy the cyclic consistency requirement: $F(G(x)) \approx x$. This equation indicates that the image \textit{x} in the \textit{X} domain can be almost completely converted back to the original \textit{x} picture after it undergoes mappings from \textit{G} to \textit{F} and then back to \textit{G}. With such cyclic consistency mapping, the model can avoid the problem of converting all the images in the \textit{X} domain to the same image in the \textit{Y} domain. Similarly, for the mapping from \textit{Y} to \textit{X}, it is also necessary to satisfy $G(F(y)) \approx y$ after learning two mappings, so that the discriminator $D_Y$ cannot recognize whether the image \textit{G(x)} is real data or generated data. As a result, the cycle of training continues until the generated image achieves the desired conversion effect.In the following,we would introduce the architectures of generators and discriminators in details.

\subsubsection{Generator Architecture}

This algorithm introduces the U-Net \cite{778045990} in the field of image segmentation to address the problem of spatial information and semantic feature loss in GAN generators. To learn more detailed information at multiple scales, this algorithm incorporates the residual block structure of the original GAN generator network at the jump junctions of the U-Net, enabling feature extraction operations to be realized on multiple branches of different scales.

The encoder part of the U-Net employs several convolutional layers for feature extraction, and its main feature is the inclusion of a large number of skip connections in the upsampling module, which passes the low-level feature information of the image to the layer with higher resolution. In the style conversion task, since the source domain image has many similar low-level features to the converted output image, skip connections can be utilized to connect the encoder in the U-Net network model to the corresponding features of the decoder, thereby helping the decoder to reuse relevant image features in the encoder. In the decoding process, the feature information extracted from the encoder can be utilized to accumulate more spatial information, effectively reducing the loss of spatial and semantic feature information, while retaining the detail information obtained from sampling different resolution feature maps. The decoder part consists of an Deconvolutional network, which combines the extracted features with the spatial semantic information obtained from skip connections to obtain the output segmentation map.

However, the U-Net network structure has some drawbacks. Since skip connections do not change the output of subsequent layers once key features have been learned in the first few convolutional layers, only the first few convolutional layers are functional within the encoder in this case. Therefore, although the U-Net network retains the spatial semantic information of the feature map, it may suffer from the problem of losing the detailed feature information of the image.

To address the above problem, we have improved the skip connections of U-Net. By appropriately adding residual block structure at the skip connection, the network can be made to learn richer detailed features. The improved structure of the generator is shown in Table 2.
\begin{table}[ht]
\caption{Generator Architecture\label{tab:tabel2}}
\centering
\resizebox{\linewidth}{!}{
\begin{tabular}{@{}c|c@{}}
\toprule
Layer name              & Net                                                                      \\ \midrule
\multirow{3}{*}{Layer1} & 3×3,16,stride 1,Normalization,ReLU                                       \\ \cmidrule(l){2-2} 
                        & Residual block1×2({[}3×3,16,stride 1,Normalization,ReLU{]}×3)            \\ \cmidrule(l){2-2} 
                        & Maxpooling                                                               \\ \midrule
\multirow{3}{*}{Layer2} & 3×3,32,stride 1,Normalization,ReLU                                       \\ \cmidrule(l){2-2} 
                        & Residual block2×2({[}3×3,32,stride 1,Normalization,ReLU{]}×3)            \\ \cmidrule(l){2-2} 
                        & Maxpooling                                                               \\ \midrule
\multirow{3}{*}{Layer3} & 3×3,64,stride 1,Normalization,ReLU                                       \\ \cmidrule(l){2-2} 
                        & Residual block3×2({[}3×3,64,stride 1,Normalization,ReLU{]}×3)            \\ \cmidrule(l){2-2} 
                        & Maxpooling                                                               \\ \midrule
\multirow{3}{*}{Layer4} & 3×3,128,stride 1,Normalization,ReLU                                      \\ \cmidrule(l){2-2} 
                        & Residual block4×2({[}3×3,128,stride 1,Normalization,ReLU{]}×3)           \\ \cmidrule(l){2-2} 
                        & Maxpooling                                                               \\ \midrule
\multirow{3}{*}{Layer5} & 3×3,16,stride 1,Normalization,ReLU                                       \\ \cmidrule(l){2-2} 
                        & Residual block1×2({[}3×3,16,stride 1,Normalization,ReLU{]}×3)            \\ \cmidrule(l){2-2} 
                        & Deconvolution4({[}3×3deconvolution,128,stride 1,Normalization,ReLU{]}×3) \\ \bottomrule
\end{tabular}}
\end{table}

\subsubsection{Discriminator Architecture}

The discriminator network is a PatchGAN network that utilizes Dilated Convolution. PatchGAN \cite{article123} is a Markovianity-based discriminator that can effectively determine the local information of the input image. Specifically, PatchGAN divides the input image into multiple regions of N×N size, binary classifies each region, and outputs an N×N matrix, where each element represents the probability that the region is judged as true or false. This matrix is then transformed into the final discriminative result through convolutional operations. Although PatchGAN can make a good judgment on the input image, it tends to limit the network's ability to acquire information on the entire image.

\begin{table}[ht]
\caption{Discriminator Architecture\label{tab:tabel3}}
\centering
\resizebox{\linewidth}{!}{
\begin{tabular}{@{}c|c@{}}
\toprule
Layer name                          & Net                                           \\ \midrule
\multirow{5}{*}{Feature Extraction} & 3×3Atrous Conv,32,stride 2,Normalization,ReLU \\ \cmidrule(l){2-2} 
                                    & 3×3Atrous Conv,32,stride 2,Normalization,ReLU \\ \cmidrule(l){2-2} 
                                    & 3×3Atrous Conv,32,stride 2,Normalization,ReLU \\ \cmidrule(l){2-2} 
                                    & 3×3Atrous Conv,32,stride 2,Normalization,ReLU \\ \cmidrule(l){2-2} 
                                    & 3×3Atrous Conv,32,stride 2,Normalization,ReLU \\ \midrule
Confidence output                   & Sigmoid                                       \\ \bottomrule
\end{tabular}}
\end{table}

To address this issue, the discriminator introduces Dilated Convolution\cite{inproceedings}. Dilated Convolution is a specialized convolution operation that can control the distance between convolutional windows by adjusting the dilation rate parameter, thereby obtaining a larger field of perception. The larger the dilation rate, the larger the convolutional window and the larger the field of perception, with the size of the field of perception increasing exponentially with the increase of the dilation rate. The number of parameters and the size of the output feature map remain constant. By utilizing dilated convolution instead of ordinary convolutional layers, the improved model in this paper can better capture global information in the image, effectively improving the discriminator's ability and performance. The structure of the improved discriminator is shown in Table 3.

\subsubsection{Loss Function}
The loss function consists of three components and is calculated as follows
\begin{equation}
\label{deqn_ex1a9}
 Loss=\ {Loss}_{GAN}+\lambda{Loss}_{cycle}+{Loss}_{identity}
\end{equation}
where ${Loss}_{GAN}$ \cite{article2323} is the adversarial loss between the distribution of the generated image and the data distribution in the target domain, which ensures that the generator and the discriminator optimize each other, and thus ensures that the generator produces a more realistic picture; ${Loss}_{cycle}$ is the cyclic consistency loss, which is computed by using the L1-norm, and is used to avoid the learned mappings \textit{G} and \textit{F} from contradicting each other, and to ensure that the output image of the generator and the corresponding input image are only different in style domain, while the content is the same;the $\lambda$ weight of the cyclic consistency loss is positively correlated with the image texture complexity; ${Loss}_{identity}$ is used to avoid the generator modifying the tones in the original image.
\begin{itemize}
\item{The adversarial loss is calculated as
%\begin{flalign}
%\begin{split}
%\end{split}
%\label{deqn_ex1a}
%{Loss}_{GAN}\ =\ L_{GAN}(G,D_Y,X,Y)\ + \\
%\ L_{GAN}(F,D_X,X,Y)=\ \mathbb{E}_{y~p_{data}}(y)[logD_Y(y)] \\
%+ Ex~pdata(x)[log(1-DY(G(x))] +\\
%Ex~pdata(x)[log(D_X(x)] + Ey~pdata(y)[log(1-D_X(F(y))]
%\end{equation}
%}
\begin{equation}
\begin{aligned}
{Loss}_{GAN}=L_{GAN}&(G,D_Y,X,Y)\ \\
+L_{GAN}(F,D_X,X,Y)=&\mathbb{E}_{y~p_{data}}(y)[logD_Y(y)]\\
+ \mathbb{E}_{x~p_{data}}(x)&[log(1-D_Y(G(x))]\\
+ \mathbb{E}_{x~p_{data}}(x)&[log(D_X(x)]\\
+ \mathbb{E}_{y~p_{data}}(y)&[log(1-D_X(F(y))].
\end{aligned}
\end{equation}
}
\item{The cyclic consistency loss is calculated as
\begin{equation}
\begin{aligned}
{Loss}_{cycle}\ =\mathbb{E}_{x~p_{data}}(x)[||F(G(x))-x{||}_1]\\
+ \mathbb{E}_{y~p_{data}}(y)[||G(F(y))-y||_1].
\end{aligned}
\end{equation}

The ultimate goal of $Minimize{\ Loss}_{cycle}$ is to make $F(G(x))\ =\ x$ as well as $G(F(y))\ =\ y$. In the case of $F(G(x))\ =\ x$, for example, the prerequisite to satisfy this goal is that the content of the generated $G\left(x\right)$ image is as close to \textit{x} as possible in terms of style close to the domain of \textit{Y}. Otherwise, it is impossible for the generator \textit{F} cannot possibly restore \textit{G(x)}, which is unconnected to x. Thus, as mentioned above, Minimize ${Loss}_{cycle}$ ensures that the content of the images produced by the generator remains roughly the same.
}
\item{ The Identity\ Loss function is calculated as
\begin{equation}
\begin{aligned}
{Loss}_{identity\ }=\ \mathbb{E}_{y~p_{data}}(y)[||D_Y(G(y))-y{||}_1]\\
+\mathbb{E}_{x~p_{data}}(x)[||D_X(G(x))-x||_1].
\end{aligned}
\end{equation}
}
\end{itemize}

\section{simulation and analysis}
\subsection{Experimental Settings}
Two publicly available datasets were utilised for this experiment: Pascal VOC 2012 \cite{article5678} and Cityscapes \cite{7780719}. These datasets were sourced from the Computer Vision Group and Mercedes-Benz Autonomous Driving Lab at the University of Oxford, UK, Max Planck Institute, and Technical University of Darmstadt, respectively. The training and testing environment for the experiments were CUDA 10.2, the programming language was Python, and the deep learning framework used was PyTorch 2.0.1.

During the semantic segmentation training process, the initial learning rate was 5e-4, the learning rate was updated using the "STEP" method, the optimization algorithm was Adam \cite{article9865}.And the number of training epochs was 300. The training set images were utilised to train the model, and the parameters obtained in the last iteration were selected and tested on the validation set. In addition, the initial learning rate for semantic restoration was 2e-4 and the same "STEP" method was used to update the learning rate. The algorithm employed Adam for parameter optimization, and the number of training epochs was set to 300.

\subsection{Analysis of Experimental Results}
\subsubsection{Convergence Experiment}
To evaluate the convergence of the semantic segmentation model proposed in this paper as well as the impact of transfer learning on network training, convergence analysis experiments are conducted in this section, and the result is shown in Fig.11.

As can be seen in Figure 11, during the training and validation process, as the number of iterations increases, the loss value of the model on both the training set and the validation set is gradually reduced, and the model is continuously optimized, indicating that both have achieved convergence. Additionally, the convergence speed of the model is also faster, indicating that the model has good stability and generalization ability. Furthermore, through comparative analysis, it can be observed that the convergence speed of the model is faster under the condition of transfer learning \cite{8237334}, and the loss values are both significantly lower than the case without transfer learning, highlighting the important application value of transfer learning in the task of image semantic segmentation.
\begin{figure}[ht]
\centering
\includegraphics[width=3.4in]{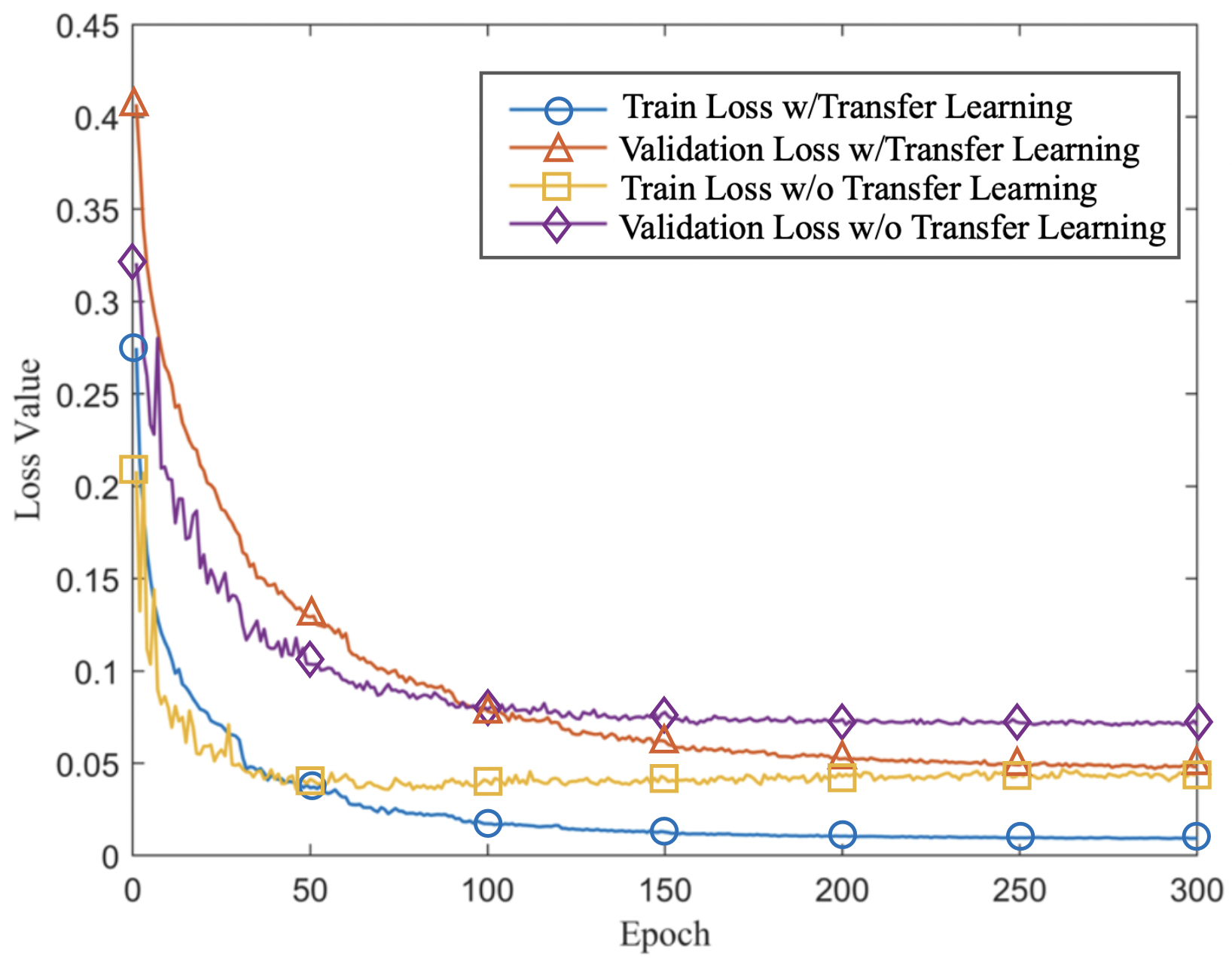}
\caption{The Convergence of the Semantic Segmentation Network.}
\label{fig_11}
\end{figure}
\subsubsection{Semantic Segmentation Compression Performance Testing}
In order to evaluate the compression effectiveness of semantic segmentation on images, this section compares the compression ratios of images processed using classical compression algorithms JPEG \cite{125072}, WebP \cite{article64278}, and the method proposed in this paper, as shown in Fig. 12.

It can be observed that the semantic images obtained by applying the image semantic segmentation algorithm proposed in this paper are able to achieve significant compression of the data volume. Under the same experimental conditions, the semantic segmentation algorithm is able to achieve more effective compression than the image compression algorithms such as JPEG and WebP by extracting semantic information from the image. This is because the image semantic segmentation algorithm aims to identify and extract the key semantic information in an image, thereby reducing the amount of data to be transmitted while retaining important information. This semantic-based compression method is more effective than the traditional pixel-based compression method because it better understands and captures the image content, thus achieving higher compression ratios.

\begin{figure}[ht]
\centering
\includegraphics[width=3.4in]{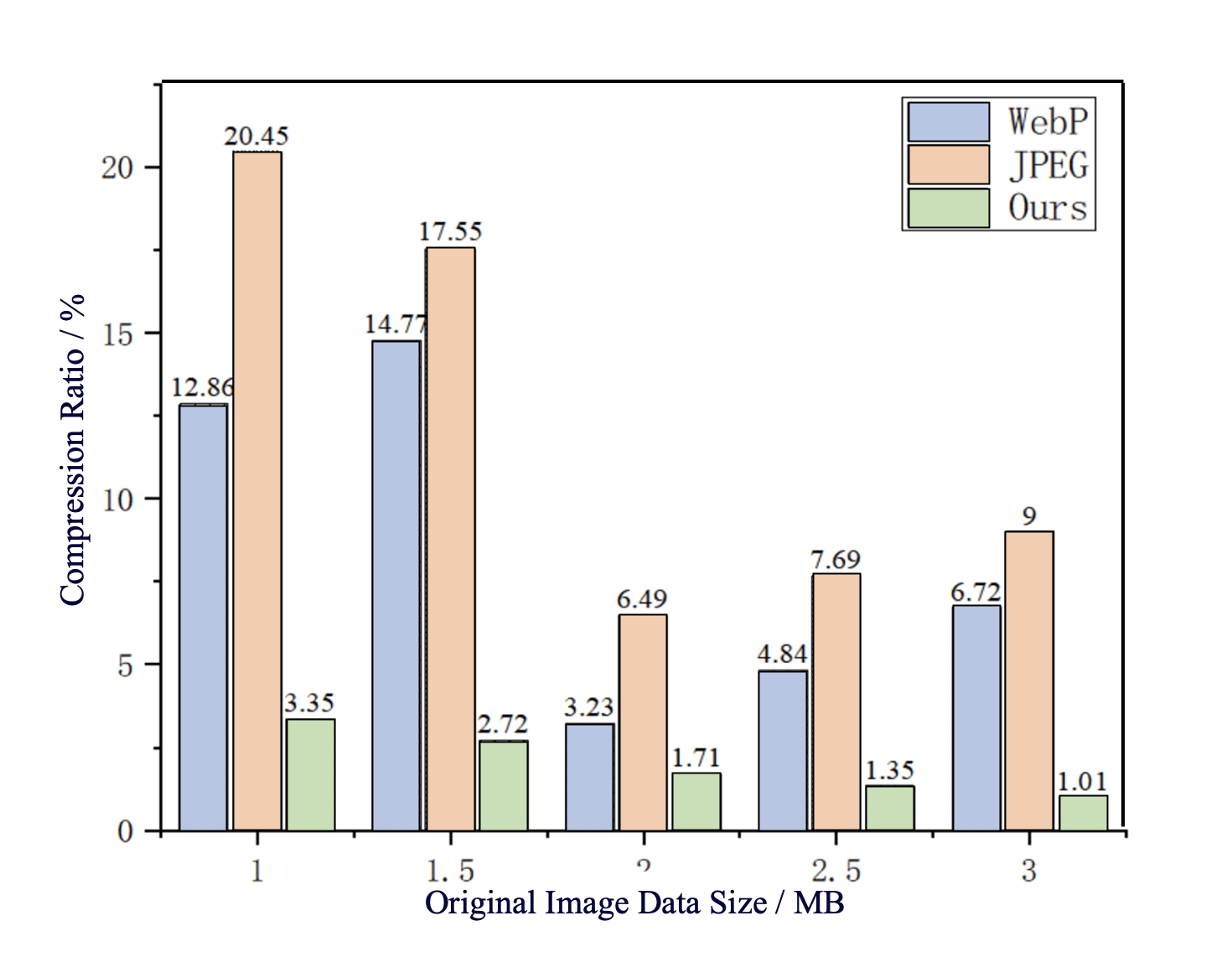}
\caption{Comparison of Algorithm Compression Performance.}
\label{fig_12}
\end{figure}
In addition, the compression ratio of the algorithm gradually decreases as the amount of original image data increases, indicating that the algorithm maintains good compression when dealing with high-resolution, high-data-volume images. This characteristic gives the algorithm a significant advantage in processing large image data, as it can effectively reduce the data volume and further improve the compression efficiency while ensuring image quality.

Therefore, applying the image semantic segmentation algorithm at the transmitter side of the AIoT system can effectively reduce transmission delay, improve transmission efficiency, and also ensure the quality of the image. This will aid the application and development of AIoT systems in real-time transmission, big data analysis, and other fields.

\subsubsection{Semantic Segmentation Accuracy Testing}
This section presents the mIoU values for each image semantic segmentation algorithm tested on the Cityscapes dataset.

\begin{table}[ht]
\caption{Segmentation algorithm accuracy testing\label{tab:tabel4}}
\centering
\begin{tabular}{@{}clll|clll@{}}
\toprule
\multicolumn{4}{c|}{Model}      & \multicolumn{4}{c}{mIoU} \\ \midrule
\multicolumn{4}{c|}{U-Net}      & \multicolumn{4}{c}{0.74} \\ \midrule
\multicolumn{4}{c|}{Hrnet}      & \multicolumn{4}{c}{0.81} \\ \midrule
\multicolumn{4}{c|}{DeepLabV3+} & \multicolumn{4}{c}{0.83} \\ \midrule
\multicolumn{4}{c|}{Ours}       & \multicolumn{4}{c}{0.86} \\ \bottomrule
\end{tabular}
\end{table}
The mIoU is a commonly used measure for evaluating the accuracy of semantic segmentation models, with a value closer to 1 indicating higher segmentation accuracy of the model. Based on the data in Table 4, the U-Net model achieved an mIoU of 0.74, which indicates lower segmentation accuracy compared to other models. This may be due to the fact that the U-Net model is susceptible to interference from background and noise when processing complex images, resulting in poor segmentation performance. HrNet \cite{9052469} and DeepLabV3+ obtained mIoU metrics of 0.81 and 0.83, respectively, and their segmentation accuracies were improved compared to U-Net. However, our proposed model achieved the highest mIoU of 0.86, indicating higher segmentation accuracy in semantic segmentation tasks.

\subsubsection{Semantic Segmentation Ablation Experiment}
To verify the effectiveness of introducing the Attention Mechanism in the Pyramid Module and training the Dice Loss and Focal Loss in the network, ablation experiments are designed in this section to demonstrate the effect of the relevant modules. The Attention Block represents adding the Attention Mechanism to the model, and Loss represents adding the Dice Loss and Focal Loss to the model. The results of the ablation experiments are shown in Table 5.
%table 5
\begin{table}[ht]
\caption{Results of ablation experiments \label{tab:tabel5}}
\centering
\begin{tabular}{@{}cccc@{}}
\toprule
Attention Block        & Loss                   & mIoU(\%)                   & mPA(\%) \\ \midrule
\multicolumn{1}{c|}{×} & \multicolumn{1}{c|}{×} & \multicolumn{1}{c|}{81.41} & 89.62   \\ \midrule
\multicolumn{1}{c|}{×} & \multicolumn{1}{c|}{\checkmark} & \multicolumn{1}{c|}{82.32} & 91.02   \\ \midrule
\multicolumn{1}{c|}{\checkmark} & \multicolumn{1}{c|}{×} & \multicolumn{1}{c|}{84.11} & 90.22   \\ \midrule
\multicolumn{1}{c|}{\checkmark} & \multicolumn{1}{c|}{\checkmark} & \multicolumn{1}{c|}{85.95} & 92.54   \\ \bottomrule
\end{tabular}
\end{table}

Based on the experimental results, we can draw the following conclusions:
\begin{itemize}
\item{Both the attention mechanism module and the associated loss function play important roles in improving image segmentation accuracy.}
\item{Introduction of the correlation loss function alone without the attention mechanism module improves segmentation accuracy by 0.91\%. This indicates that the Dice Loss and Focal Loss can improve segmentation accuracy to some extent.}
\item{When only the attention module is added and the Dice Loss and Focal Loss are not included, segmentation accuracy improves by 2.7\%. This indicates that the attention mechanism module can significantly improve segmentation accuracy.}
\item{Introduction of both the correlation loss function and the attention mechanism module together improves image segmentation accuracy by 4.54\%. This indicates that these two modules can complement each other to improve segmentation accuracy together.}
\end{itemize}

The experimental results demonstrate the importance of different modules in the network model and effectiveness of improving image segmentation accuracy. They also provide valuable references and insights for subsequent research.

\subsubsection{Experiments on Quantitative Modeling of Semantic Segmentation}
Table 6 compares the mIoU before and after quantization of the model, as well as the time required to process a 2MB-sized image.

By quantizing the model, although we observe a decrease of 0.07 mIoU values, the processing speed increases by 32.10\%. This indicates that although the accuracy of the model has decreased, the increase in processing speed better meets the demand of low-latency in some scenarios.

In some application scenarios, low-latency may be more important than high-accuracy. For example, in real-time video stream processing, it is more important to provide results as quickly as possible than to provide 100\% accurate results. Therefore, the needs of these specific scenarios can be better met by quantizing the model to increase the processing speed while slightly decreasing the accuracy.
\begin{table}[ht]
\caption{Quantitative modeling experimental results \label{tab:tabel6}}
\centering
\begin{tabular}{@{}c|c|c|c@{}}
\toprule
Model         & MIoU & Running time & QAR     \\ \midrule
Ours(w/ QAT)  & 0.86 & 0.324        & -       \\ \midrule
Ours(w/o QAT) & 0.79 & 0.223        & 32.10\% \\ \bottomrule
\end{tabular}
\end{table}
\begin{figure}[ht]
\centering
\includegraphics[width=3.3in]{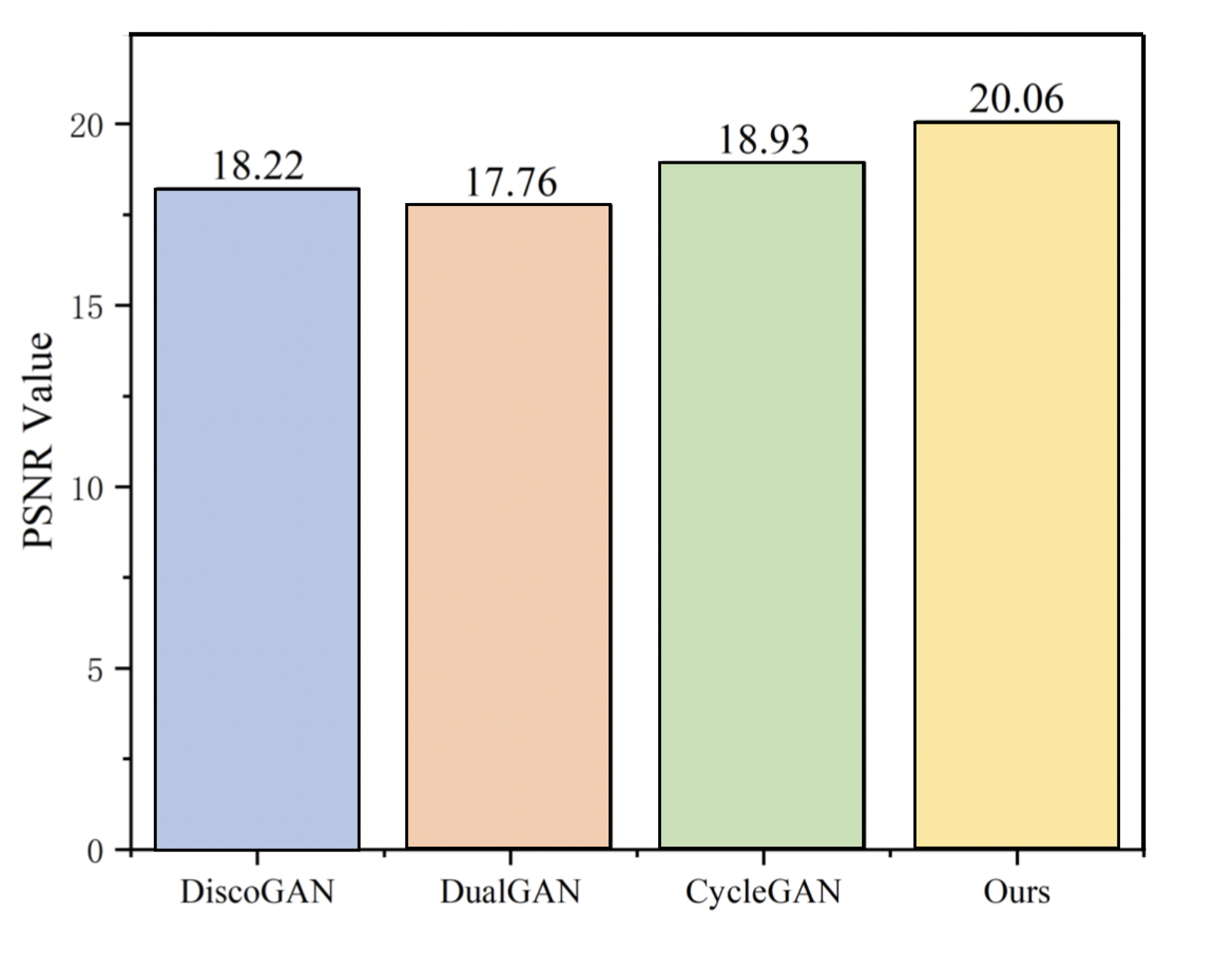}
\caption{Comparison of PSNR Values of Different Restoration Algorithms.}
\label{fig_13}
\end{figure}

\begin{figure}[ht]
\centering
\includegraphics[width=3.2in]{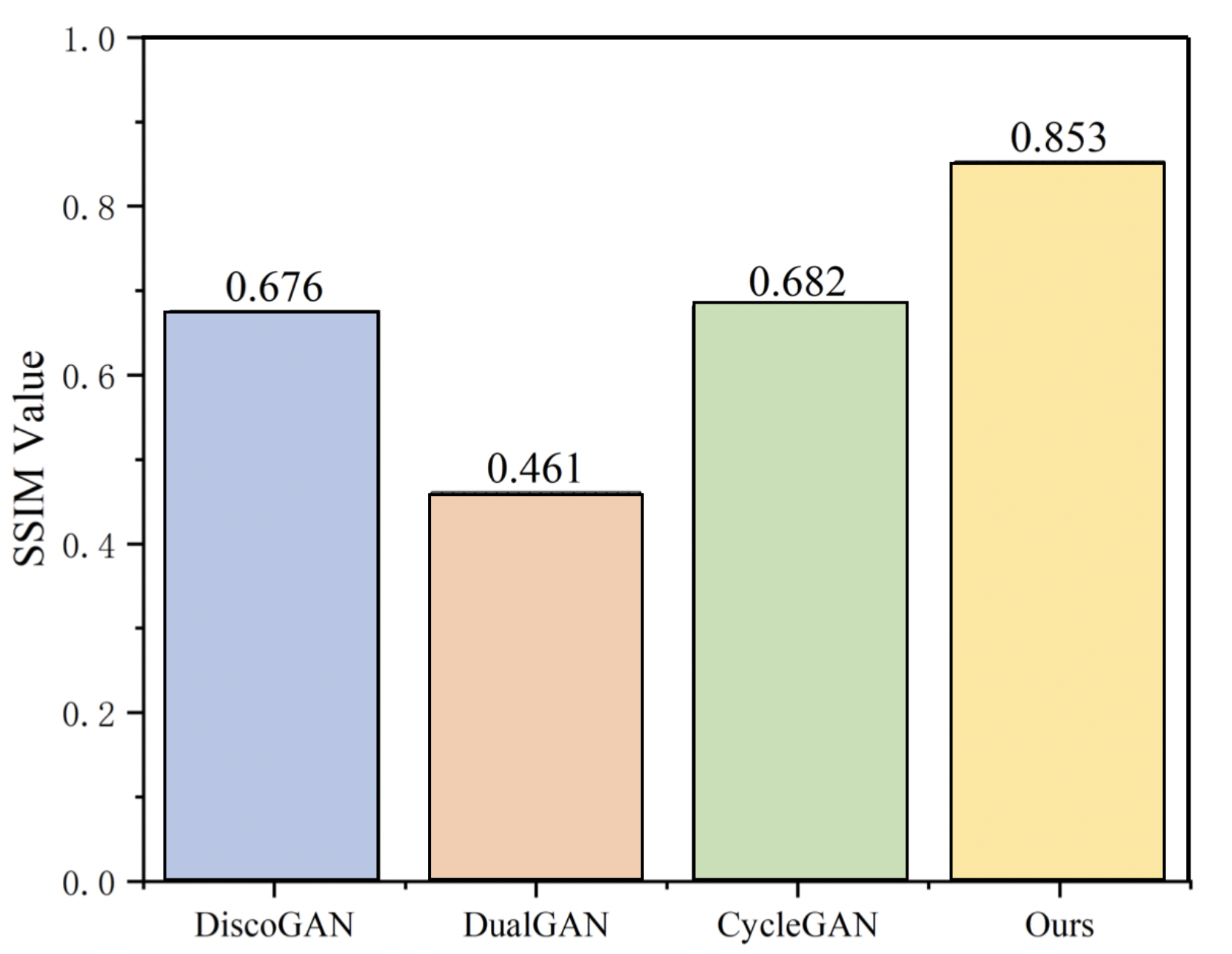}
\caption{Comparison of SSIM Values of Different Restoration Algorithms.}
\label{fig_14}
\end{figure}
\subsubsection{Semantic Restoration Performance Experiments}
This algorithm is designed to transform a semantic image into a real image with detailed texture visible to the naked eye. To compare this algorithm with classical algorithms DiscoGAN \cite{article67369}, DualGAN \cite{8237572}, CycleGAN \cite{article38738}, and other related algorithms, two metrics PSNR \cite{5596999}and SSIM \cite{5596999} are employed, respectively. The results are shown in Figs. 13 and 14.

It can be observed that by utilizing the algorithm proposed in this paper, improvements are observed in both the structural similarity and peak signal-to-noise ratio metrics. Among them, the average SSIM score reaches 0.853 and the average PSNR score reaches 20.64. Compared to other classical algorithms, this algorithm demonstrates superior performance, indicating that the restored images possess higher quality.

\subsubsection{Total Latency Performance Testing}
The aim of this experiment is to test and compare the total delay between the original image communication and the deep image semantic based image communication under the same spectral resource conditions. The system configuration for this experiment is shown in Fig.15, which mainly consists of three components: the Nvidia Jetson AGX Orin, the USRP NI2901, and the touch screen. In this system, the Nvidia Jetson AGX Orin acts as the main control device responsible for controlling the overall experimental process; the USRP NI2901 is responsible for implementing the transceiver function of wireless signals; and the touch screen is used for user interaction and data display.

The operating environment of the entire demonstration system is as follows: the operating system is Ubuntu 20.04, and GNU Radio 3.9 is used for the design and debugging of the wireless communication system. The data carrier frequency is set as 2.1 GHz, the ACK carrier frequency is 900MHz, the antenna gain is 40dB, and GMSK modulation is used.

\begin{figure}[ht]
\centering
\includegraphics[width=3.4in]{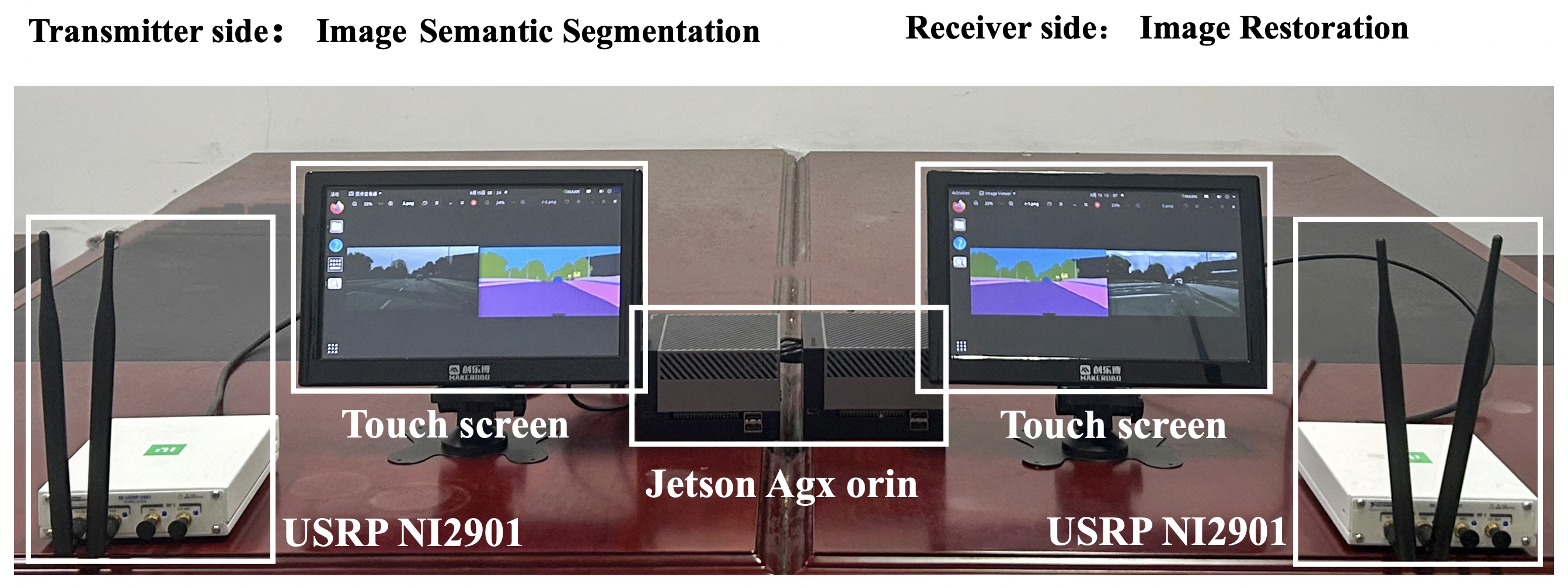}
\caption{Schematic Diagram of Experimental DEMO construction.}
\label{fig_15}
\end{figure}
Fig.16 presents the total delay of the original image transmission and proposed model in this paper, for different data sizes and transmission rates. Herein, the total delay of the original image transmission refers to the total time required for transmitting the original image, while the total delay of proposed model includes the time for semantic segmentation of the original image, the time for semantic image transmission, and the time for semantic restoration.

\begin{figure}[ht]
\centering
\includegraphics[width=3.4in]{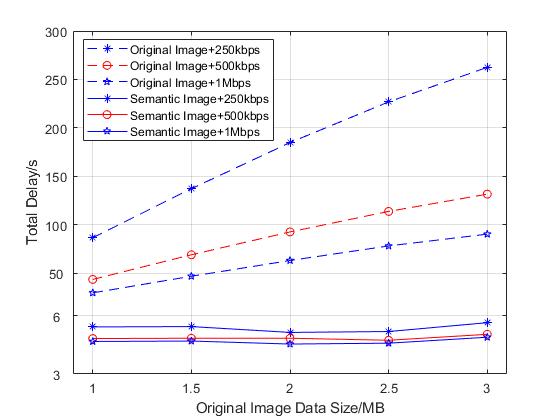}
\caption{Comparison of Total Delay Under Different Communication Methods.}
\label{fig_16}
\end{figure}
As Fig.16 clearly indicates, the proposed method displays a significant advantage in terms of total delay performance compared to the the original image transmission. Data analysis results demonstrate that proposed method reduces the total delay by an average of 95.26\% compared to the original image transmission. This significant advantage significantly reduces the time required for transmission, thereby enhancing the transmission efficiency of the communication system.

\section{conclusion}
This paper has investigated several limitations in processing image data, including the trade-off between semantic information extraction accuracy and compression performance, the lack of consideration for the impact of semantic restoration on real images, and the cumbersome network model, and proposed a novel deep image semantic communication system model. We have demonstrated the model can achieve significant compression effects at the transmitter by segmenting the original image into high-precision semantic images and reducing the storage and computational requirements; and convert the semantic image into a real scene image, at the receiver and ensure the sustainability of subsequent tasks and image visibility. For the future work, we will study performance issues at low signal-to-noise ratios and generalization capability in more complex channels to expand the application scope and performance of image semantic communication models. Additionally, we will look into it how image semantic communication models can be effectively compressed to further reduce computational overhead at both the transmitter and receiver sides, and reduce latency to ensure real-time performance of semantic communication systems.

\bibliography{zy.bib}
\bibliographystyle{IEEEtran}

\vspace{11pt}

\vfill

\end{document}